\documentclass{article}

\usepackage[main, final]{neurips_2026}

\usepackage[utf8]{inputenc} % allow utf-8 input
\usepackage[T1]{fontenc}    % use 8-bit T1 fonts
\usepackage{hyperref}       % hyperlinks
\usepackage{url}            % simple URL typesetting
\usepackage{booktabs}       % professional-quality tables
\usepackage{amsfonts}       % blackboard math symbols
\usepackage{nicefrac}       % compact symbols for 1/2, etc.
\usepackage{microtype}      % microtypography
\usepackage[table]{xcolor}  % colors (table option for \rowcolor/\cellcolor)

\usepackage{multirow}
\usepackage{enumitem}
\usepackage{makecell}
\usepackage{amsmath}
\usepackage{amssymb}
\usepackage{graphicx}
\usepackage{subcaption}
\usepackage{gensymb}
\definecolor{labelgray}{gray}{0.5}
\usepackage{wrapfig}
\usepackage{tabularx}
\usepackage{float}
\usepackage[most]{tcolorbox}
\hypersetup{
    hidelinks
}
\usepackage{fontawesome5}
\newcommand{\mIoU}{mIoU~($\uparrow$)}
\newcommand{\mAcc}{mAcc~($\uparrow$)}

\title{RelationVGGT: Visual Geometry Transformers for 3D Spatial Relation Segmentation
}

\author{
    \begin{tabular}{c}
    \href{https://minsu1206.github.io/}{Minsu Kim}$^{1}$
    \hspace{1.6em}
    \href{https://jaesung-choe.github.io/}{Jaesung Choe}$^{2}$
    \hspace{1.6em}
    \href{https://gibsonlee01.github.io/}{Jiwoo Lee}$^{1}$
    \\[0.05em]
    \href{https://vllab.ee.ntu.edu.tw/ycwang.html}{Yu-Chiang Frank Wang}$^{2}$
    \hspace{1.8em}
    \href{https://sites.google.com/site/seonjookim/}{Seon Joo Kim}$^{1}$
    \\[0.55em]
    $^{1}$Yonsei University
    \hspace{2em}
    $^{2}$NVIDIA
    \end{tabular}
}

\begin{document}

\maketitle
\vspace{-1em}

{\centering\small
\href{https://relationvggt.github.io/}{%
    {\faGlobe\enspace\textbf{Project Page}}%
}\par}

\vspace{1.2em}
\begin{abstract}
    Recent advances in 3D reconstruction have progressed from per-scene optimization to feed-forward inference, and semantic scene understanding has followed suit—yet existing methods remain confined to object-centric perception, neglecting spatial relations between objects.
We formulate \emph{3D spatial relation segmentation} in a feed-forward, pose-free multi-view setting: given a visually specified subject and a relational text query, the model segments the target across views without receiving its category name.
To this end, we propose RelationVGGT, a novel feed-forward framework that integrates semantic features from a visual foundation model with geometry-aware representations from a 3D geometry foundation model and leverages a relation transformer for subject-conditioned, cross-view relation prediction—requiring neither per-scene optimization nor known camera poses. 
We additionally provide a fully automated annotation pipeline built on ScanNet++ with VLMs and LLMs, enabling scalable training data generation for this new task.
\end{abstract}

\section{Introduction}

Recent years have witnessed remarkable advancements in 3D reconstruction and neural rendering.
Pioneering representations such as Neural Radiance Fields (NeRF)~\cite{mildenhall2021nerf} and 3D Gaussian Splatting (3DGS)~\cite{3dgs} have revolutionized scene-optimized view synthesis and geometric modeling.
More recently, visual geometry models such as DUSt3R~\cite{wang2024dust3r} and VGGT~\cite{wang2025vggt} have demonstrated that dense 3D geometry can be predicted directly from multi-view images without camera pose priors or Structure-from-Motion pipelines, fundamentally shifting how multi-view geometry can be inferred.

This shift has not been limited to reconstruction: semantic scene understanding has followed the same trajectory.
Methods such as LangSplat~\cite{langsplat} and OpenGS~\cite{wu2024opengaussian} embed vision-language features into scene-optimized 3D primitives, enabling open-vocabulary 3D scene understanding and zero-shot object segmentation from simple textual queries like `chair' or `table'.
Beyond per-scene methods, feed-forward approaches like PanSt3R~\cite{zust2025panst3r} and Uni3R~\cite{sun2025uni3r} integrate geometry foundation model~\cite{wang2024dust3r, wang2025vggt, wang2025pi3}  priors to achieve multi-view consistent semantic segmentation, further bypassing the need for per-scene optimization and posed images.

While these previous studies have made significant strides, they dominantly focus on object-centric scene understanding—essentially treating environments as mere collections of isolated entities. 
However, holistic scene perception requires a deeper understanding of how objects interact.
We understand an environment by recognizing how objects are spatially arranged and related to one another---a monitor \emph{resting on} a desk, a mug \emph{next to} a keyboard.
Such relational reasoning underpins higher-order tasks such as spatial question answering and is a core capability that embodied agents must acquire to act in 3D environments.
This motivates us to study relational 3D reasoning beyond object-level perception, particularly in a generalizable multi-view setting.

To this end, we formulate the task of \emph{3D spatial relation segmentation} in a feed-forward, pose-free setting, where the goal is to localize, across scene views, the object satisfying a spatial relationship determined from the underlying 3D structure.
Given multi-view images, a subject mask in a reference view, and a relational text query (e.g, ``resting on''), the model must predict segmentation masks of the target object consistently across all target views.
This formulation goes beyond conventional 3D segmentation by requiring semantic identity, directional spatial relation reasoning, and cross-view geometric consistency.

Existing approaches offer two natural starting points for this formulation.
RelationField~\cite{koch2025relationfield} learns inter-object relationships on top of radiance fields, but relies on scene-specific optimization with calibrated camera poses, limiting its scalability and generalizability to unseen scenes.
Alternatively, video-language models perform multi-view relational reasoning, but without a 3D scene representation, they struggle to maintain object correspondences under large viewpoint changes.

Motivated by these limitations, we propose \textbf{RelationVGGT}, a feed-forward framework for 3D spatial relation segmentation that requires neither per-scene optimization nor known camera poses.
Our core idea is to reformulate the volumetric relation search of prior work as a per-pixel relation prediction problem on the target-view image grid, conditioned on the reference subject.
To this end, we fuse semantic features from DINOv2~\cite{oquab2023dinov2} with geometry-aware representations from the Pi3~\cite{wang2025pi3} encoder, and employ a relation transformer that alternates subject-conditioned cross-attention with multi-view self-attention to produce language-aligned relation features for open-vocabulary segmentation.
A critical challenge is the scarcity of training data with multi-view consistent relation annotations.
To address this, we build a fully automated data generation pipeline on top of ScanNet++~\cite{yeshwanth2023scannetpp}, exploiting instance-level labels and VLM/LLM-based relation extraction to produce scalable training data.

In summary, our contributions are threefold.
First, we formulate \emph{3D spatial relation segmentation} in a feed-forward, pose-free multi-view setting, requiring models to localize target objects defined by spatial relations consistently across views. 
Second, we propose RelationVGGT, a feed-forward architecture that fuses 2D semantic and 3D geometry-aware features and reformulates volumetric relation search as per-pixel, subject-conditioned relation prediction.
Third, we provide a fully automated relation annotation pipeline leveraging VLMs and LLMs on top of ScanNet++, enabling scalable training data generation.
\section{Related Works}
\label{sec:related-work}

\paragraph{3D Scene Understanding with Foundation Models.}

Recent advances in 3D scene understanding span scene-optimized representations such as NeRF~\cite{mildenhall2021nerf} and 3DGS~\cite{3dgs}, as well as feed-forward geometry foundation models~\cite{wang2024dust3r, wang2025vggt, wang2025pi3, mast3r, fast3r, cut3r} that predict dense 3D geometry directly from multi-view images without known camera poses.
Building on these, methods such as LangSplat~\cite{langsplat} and follow-up works~\cite{wu2024opengaussian, laga, zhu2025cos3d, ilgs, zhu2025objectgs} embed vision-language features into 3D primitives for open-vocabulary segmentation, while feed-forward approaches~\cite{zust2025panst3r, sun2025uni3r} achieve multi-view consistent semantics without per-scene optimization.
However, these methods remain object-centric, localizing individual entities without reasoning about inter-object spatial relations.
Our work builds on these foundations, fusing DINOv2~\cite{oquab2023dinov2} semantics with geometry-aware representations to enable relational reasoning beyond object-level perception.

\paragraph{3D Spatial Relation and Scene Graph.}
Understanding relationships between objects has traditionally been formulated as Scene Graph Generation (SGG), where nodes represent objects and edges encode pairwise predicates.
While 2D SGG methods~\cite{xu2017scenegraph, zellers2018neuralmotif, tang2018learningcomposedynamictree} have been widely studied, they cannot capture the multi-view nature of 3D environments.
3D scene graph methods~\cite{wu2021SceneGraphFusion, wald20203DSSG, gu2023conceptgraphs} construct graphs from 3D point clouds and jointly predict object nodes and relation edges.
However, these approaches assume access to pre-segmented object instances and focus on classifying relations from a fixed predicate vocabulary.
More recently, RelationField~\cite{koch2025relationfield} has pioneered relational understanding in neural radiance fields by learning continuous relation features conditioned on pairs of 3D coordinates.
However, it requires scene-specific optimization with calibrated camera poses and localizes relations through queries over the reconstructed 3D representation.
In contrast, we replace 3D relation search with feed-forward per-pixel prediction over unposed multi-view images.
More broadly, we take a relation-first perspective: rather than recovering object nodes and then classifying their edges, we directly segment the target conditioned on a visually specified subject and a relation query.
% --------
This formulation is complementary to scene graph generation and could be integrated with node prediction methods to jointly recover full scene graphs in future work.

\paragraph{Visual Grounding.}
VLM-based methods such as LISA~\cite{lisa} enable open-vocabulary visual grounding from natural-language queries, while Video-LISA~\cite{video-lisa} extends this capability to video inputs.
Recent approaches such as SA2VA~\cite{yuan2025sa2va} and UniPixel~\cite{liu2025unipixel} further support dense pixel-level mask prediction for grounded targets.
Despite their strong grounding capability, these methods do not explicitly model the underlying 3D scene structure and are primarily designed for image or temporally continuous video observations, making relational grounding more challenging under sparse, wide-baseline multi-view inputs.

Closely related 3D referring segmentation methods~\cite{refersplat, concretenet, langref3dgs, mvggt} localize targets from language descriptions in 3D or multi-view representations.
Their formulation, however, differs from ours: a textual referring expression identifies the target object, potentially through its category, attributes, or relation to other objects.
In contrast, we visually specify the reference subject with a mask and provide only the relation as text; the target identity and category are not given and must be inferred from the subject--relation pair.
\section{Method}
\subsection{3D Spatial Relation Segmentation}
\paragraph{Problem Definition and Notations.}
We introduce \emph{3D spatial relation segmentation}, a task that aims to localize, across multiple views of a scene, the object satisfying a spatial relationship that can only be determined from the underlying 3D structure of the scene.
Here, spatial relations refer to relationships arising from the 3D arrangement of objects, such as support, containment, and proximity.
Consider a scene captured by a set of $N$ images $\mathcal{I}=\{I_i\}_{i=1}^{N}$.
One image $I_r$ is designated as the \emph{reference view}, and the remaining images are treated as \emph{target views}, denoted by $\mathcal{I}_{\mathrm{tgt}}=\mathcal{I}\setminus\{I_r\}$.
In the reference view, a binary mask $M_s$ indicates the \emph{subject}, i.e., the query object from which the relation originates.
A relational text query $q$ specifies the spatial relationship that the desired object should satisfy with respect to the subject.
We refer to this desired object as the \emph{target object}.
For example, if the subject mask corresponds to a monitor and the relation query is ``resting on,'' the target object is the desk supporting the monitor.
The goal is to predict, for each target view $I_t \in \mathcal{I}_{\mathrm{tgt}}$, a segmentation mask that localizes the target object from the viewpoint of $I_t$.

\paragraph{Motivation.}
A natural baseline for this task is RelationField~\cite{koch2025relationfield}, which extends a radiance field with an additional query coordinate to model relational semantics:
\begin{equation}
  g_{\theta}(\mathbf{x}, \mathbf{z}) \;\mapsto\; \mathbf{r},
  \label{eq:relationfield}
\end{equation}
where $\mathbf{x}$ is the target 3D position, $\mathbf{z}$ the query 3D position, and $\mathbf{r}$ the predicted relation feature.
The two positions are encoded via positional encodings and concatenated as input to the MLP network.
This design has two fundamental limitations.
First, because both the query and target are represented as 3D coordinates, locating the target object requires evaluating $g_\theta$ at every candidate position in the scene volume.
For example, if the subject is a monitor and the relation is ``resting on,'' the model should identify the desk supporting it—yet to do so, it must sweep through the entire 3D space, a search that is both computationally expensive and inherently dependent on an explicit volumetric reconstruction.
Second, the reliance on scene-specific positional encodings ties the learned representations to a particular coordinate system, leaving generalization to unseen scenes unexplored.

Our key observation is that this volumetric search is unnecessary.
Given the subject mask $M_s$ in the reference view, we can formulate target localization on the image grid of each target view, rather than over a dense 3D volume.
This replaces an unstructured volumetric search with a per-pixel relation prediction problem conditioned on the reference subject, without relying on a scene-specific 3D field at test time.
Building on this observation, we define a per-view \emph{relation feature map}:
\begin{equation}
  \mathcal{R}_t = \bigl\{\, \mathbf{r}_{t}(\mathbf{u}) \;\big|\; \mathbf{u} \in I_t \bigr\},
  \label{eq:relation-feature-map}
\end{equation}
where $\mathbf{r}_{t}(\mathbf{u}) \in \mathbb{R}^{D}$ encodes the spatial relationship between location $\mathbf{u}$ in target view $I_t$ and the reference subject specified by $M_s$.
By aligning this feature with the language embedding space, open-vocabulary relational queries can be evaluated through feature–text similarity, yielding dense relevance maps from which segmentation masks are derived.

Formally, our model learns a function $f_\theta$:
\begin{equation}
  f_{\theta}(I_r,\, M_s,\, \mathcal{I}_{\mathrm{tgt}},\, \mathbf{u}) \;\rightarrow\; \mathcal{R}_t,
  \label{eq:relation-prediction}
\end{equation}
which predicts the relation embedding at each target-view location conditioned on the reference subject.
Given the query embedding $\mathbf{e}(q)$, the target object is segmented by measuring compatibility between $\mathbf{r}_{t}(\mathbf{u})$ and $\mathbf{e}(q)$ over all $\mathbf{u}\in\Omega_t$.
This reformulation eliminates the need for volumetric search and scene-specific 3D optimization, while supporting generalization to novel scenes through multi-view-aware representations.

\subsection{Architecture}
\vspace{-7pt}
\begin{figure*}[t]
    \centering
    \includegraphics[width=\linewidth]{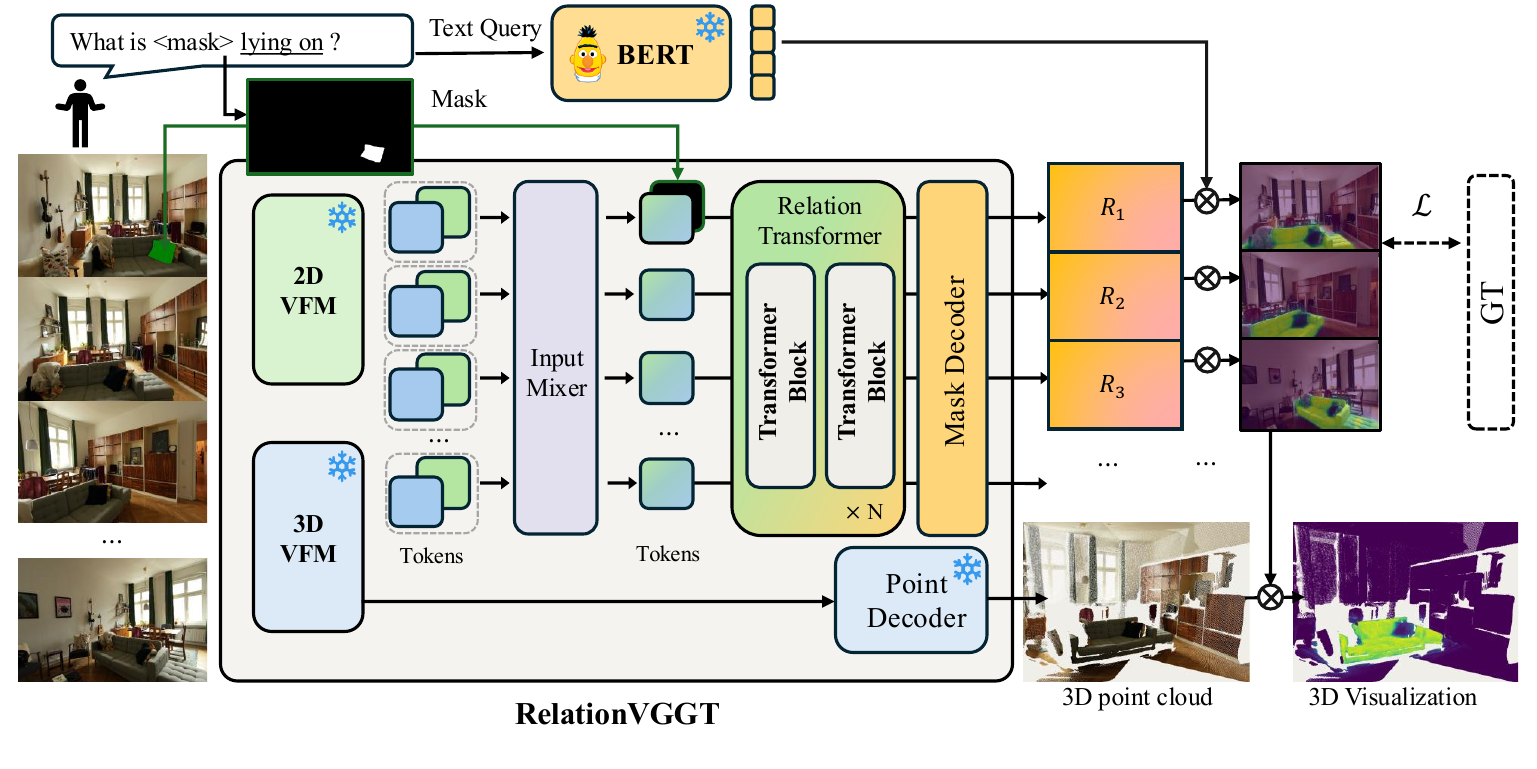}
    \vspace{-10pt}
    \caption{\textbf{Overview of RelationVGGT.}
Multi-view images are encoded by frozen 2D and 3D visual foundation models and fused via an input mixer. The subject mask conditions the relation transformer, whose output is decoded into per-view relation feature maps. The target object is segmented via feature--text similarity with the BERT-encoded relational query.}
   \label{fig:method}
\end{figure*}
\vspace{-3pt}

\label{sec:architecture}
The overall pipeline of RelationVGGT is illustrated in Fig.~\ref{fig:method}.
Given a reference image $I_r$ with a subject mask $M_s$ and target images $\mathcal{I}_{\mathrm{tgt}}$, our model extracts dense patch-level representations for all views, performs subject-conditioned relation reasoning across views, and decodes pixel-aligned relation features for open-vocabulary segmentation.

\paragraph{Feature extraction.}
\label{subsec:feature_extraction}
Relational reasoning requires both object-level semantics and spatially grounded scene structure.
Following \cite{zust2025panst3r}, we extract two complementary dense representations from the input image set $\mathcal{I}$.
For semantic cues, we use DINOv2~\cite{oquab2023dinov2} feature maps, which provide dense patch-level visual representations with strong object-level semantics.
Geometry-aware cues are obtained from the Pi3~\cite{wang2025pi3} encoder tokens before its task-specific prediction heads, preserving general multi-view geometric information.
The two feature streams are represented on a common patch grid and subsequently fused by the input mixer.

For each image $I_n \in \mathcal{I}$, the two feature streams are denoted as
\begin{equation}
E_n^{D} = \mathrm{Enc}_{D}(I_n),
\qquad
E_n^{P} = \mathrm{Enc}_{P}(\mathcal{I})_n.
\label{eq:foundation-features}
\end{equation}
We concatenate them along the channel dimension and map them to a compact joint representation using the input-mixing architecture of~\cite{zust2025panst3r}:
\begin{equation}
F_n = M_{\theta}\!\left(\left[E_n^{D};\, E_n^{P}\right]\right)
\in \mathbb{R}^{P \times d_f},
\label{eq:input-mixing}
\end{equation}
where $M_{\theta}$ is the input mixer, $[\,\cdot\,;\,\cdot\,]$ denotes channel-wise concatenation, and $P = H_p W_p$ is the number of patches per image.
% The mixer $M_{\theta}$ consists of a linear projection, a small stack of transformer blocks with RoPE positional encoding, and a final layer normalization.
% This produces joint semantic-geometric tokens
% $F_n=[\mathbf{f}_n(\mathbf{u}_1),\ldots,\mathbf{f}_n(\mathbf{u}_P)]$,
% which are passed to the subsequent relation reasoning module.
The mixer consists of a linear projection followed by a small stack of transformer blocks with RoPE~\cite{su2024rope} positional encoding.
This produces joint semantic-geometric tokens $F_n$ for each view, which are passed to the relation reasoning stage.

\paragraph{Relation prediction.}
\label{subsec:relation_prediction}
To predict a per-pixel relation feature map for each target view, the model must condition target-view representations on the reference subject.
A key design choice is how to represent the subject.
Rather than compressing it into a single pooled vector, we retain the full set of reference-view patch tokens and mask out non-subject patches, thereby preserving the subject's spatial layout and fine-grained semantics.
Concretely, let $F_r \in \mathbb{R}^{P \times d_f}$ denote the reference-view tokens and $\bar{M}_s \in \{0,1\}^{P}$ be the subject mask downsampled to the patch grid.
The subject-masked reference tokens are formed as
\begin{equation}
F_r^s
=
\bar{M}_s \odot F_r
\in \mathbb{R}^{P \times d_f},
\label{eq:masked-subject-tokens}
\end{equation}
where $\odot$ denotes element-wise multiplication, zeroing out non-subject patches.
Target-view tokens are stacked into a single sequence for joint processing:
\begin{equation}
Z^0
=
\operatorname{Concat}\!\left(\{F_t\}_{I_t \in \mathcal{I}_{\mathrm{tgt}}}\right)
\in \mathbb{R}^{(N-1)P \times d_f}.
\end{equation}
The relation transformer then refines this sequence by alternating subject-conditioned cross-attention and global self-attention:
\begin{equation}
\bar{Z}^{\ell}
=
\operatorname{CrossAttn}_{\ell}
\!\left(
Z^{\ell},\, F_r^s
\right),
\qquad
Z^{\ell+1}
=
\operatorname{SelfAttn}_{\ell}
\!\left(
\bar{Z}^{\ell}
\right).
\label{eq:relation-transformer}
\end{equation}
Cross-attention allows each target patch to attend to the subject-masked reference tokens, while self-attention enables information exchange among target patches across all views.
This makes the resulting representation both subject-conditioned and multi-view-aware.

After $L$ layers of the relation transformer, the updated tokens are reshaped back into per-view patch grids.
Since the transformer operates at patch resolution, a mask decoder upsamples the tokens into a pixel-aligned relation feature map:
\begin{equation}
\mathcal{R}_t
=
\mathrm{Dec}
\!\left(
Z_t^{L}
\right)
= \left\{\, \mathbf{r}_t(\mathbf{u}) \;\big|\; \mathbf{u} \in I_t \right\},
\label{eq:relation-map-head}
\end{equation}
where $Z_t^{L}$ denotes the final tokens corresponding to target view $I_t$, and $\mathcal{R}_t$ is the resulting relation feature map aligned with the language embedding space.
In practice, we implement the mask decoder using a DPT head~\cite{dpt}.
The target object is then localized by computing feature–text similarity with the query embedding $\mathbf{e}(q)$, as detailed in Sec.~\ref{sec:inference}.

\paragraph{3D geometry.}
Additionally, we use the Pi3~\cite{wang2025pi3} point decoder to predict a point map and confidence map for each target view.
Since $\mathcal{R}_t$ is pixel-aligned, each relation feature $\mathbf{r}_t(\mathbf{u})$ can be associated with its corresponding 3D point, yielding a confidence-aware 3D relation representation across views.

\subsection{Training}
\label{sec:training}
The spatial relationship between two objects can often be described by multiple valid expressions.
For example, the relational query ``standing on'' may equally well be expressed as ``on'' or ``supported by.''
Treating this as a single-label classification problem may penalize semantically plausible alternatives.
To address this, we supervise the same target mask with multiple compatible relational queries, treating each query independently with a sigmoid focal loss~\cite{lin2018focalloss}.
The full training objective combines this multi-query focal loss with a Dice loss for spatial mask alignment:
\begin{equation}
\mathcal{L}_{\mathrm{mask}}
=
w_{\mathrm{focal}}\,
\mathcal{L}_{\mathrm{focal}}^{\mathrm{multi\text{-}q}}
+
w_{\mathrm{dice}}\,
\mathcal{L}_{\mathrm{dice}}.
\end{equation}
% \minsu{TODO: detail appendix.}
We provide further details of the multi-query supervision and training objective in the appendix.
\subsection{Inference}
\label{sec:inference}
Given the predicted relation feature map $\mathcal{R}_t$, we compute a relevance logit $z_t(\mathbf{u};q)$ at each pixel by matching the relation feature with the text query embedding:
\begin{equation}
z_t(\mathbf{u};q)
=
\mathbf{r}_t(\mathbf{u})^\top \mathbf{e}(q).
\end{equation}
The corresponding mask score is obtained by applying sigmoid to the logits:
\begin{equation}
S_t(\mathbf{u};q)
=
\operatorname{sigmoid}
\left(
z_t(\mathbf{u};q)
\right).
\end{equation}
Unlike feature field methods that require negative-prompt normalization, we empirically find that directly applying sigmoid suffices without softmax over positive and negative text embeddings.
The final binary mask is obtained using a fixed threshold $\tau=0.3$:
\begin{equation}
\hat{M}_t(q)
=
\left\{
\mathbf{u}
\,\middle|\,
S_t(\mathbf{u};q) \ge \tau
\right\}.
\end{equation}
We use the same threshold for all experiments without additional post-processing.
\vspace{-5pt}
\begin{figure*}[t]
    \centering
    \includegraphics[width=\linewidth]{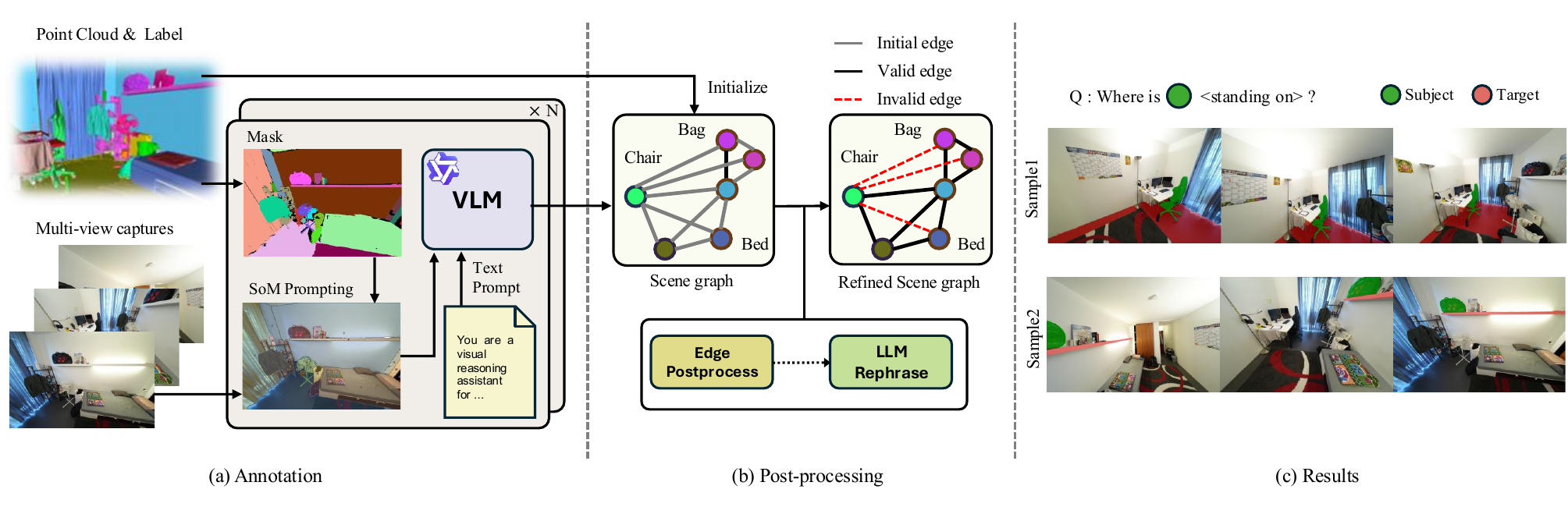}
    \vspace{-20pt}
    \caption{\textbf{Overview of the relation annotation pipeline.}
(a)~Instance masks are projected onto frames with SoM prompting to query a VLM for pairwise relations.
(b)~Proposals are aggregated into a scene graph and refined via geometry-aware postprocessing and LLM.
(c)~Example results showing multi-view consistent subject (green) and target (red) annotations.}
   \label{fig:dataset}
   \vspace{-10pt}
\end{figure*}

\subsection{Dataset}
\label{sec:dataset}
Our model requires multi-view consistent relation labels with object-level masks, but such supervision is scarce and costly to annotate manually.
We therefore build an automatic annotation pipeline on ScanNet++~\cite{yeshwanth2023scannetpp}, which provides diverse indoor scenes with 3D instance-level point annotations.
The overview of the pipeline is shown in Fig.~\ref{fig:dataset}.

For each scene, we initialize a scene graph with object instances as nodes and edges representing pairwise spatial relations, initially left empty.
We project instance masks onto posed RGB frames, apply Set-of-Mark (SoM) prompting to mark visible objects, and query a VLM to obtain frame-wise relation proposals.
These proposals are collected as candidate edges but are not yet treated as valid relations, as they must pass the following multi-view and geometric verification steps.

Since frame-wise VLM proposals are often noisy, we filter unreliable predictions by requiring agreement across multiple viewpoints.
For each object pair visible across multiple frames, the VLM independently proposes relations from each viewpoint.
We count how many frames predict each relation for a given pair.
If the count falls below $K{=}3$, the relation is rejected as insufficiently supported.
Otherwise, it is retained as a candidate edge.
Since we already know which frames each subject-target pair appears in via rasterization of 3D instance points, retained relations are propagated to all co-visible frames, completing the multi-view annotation.

We further apply a geometry-based filtering step to remove geometrically wrong predictions.
Using 3D bounding boxes and instance centroids, we reject relations between object pairs whose bounding boxes do not overlap and whose centroids are farther than a distance threshold, such as contact or support relations predicted between distant objects.

We observe that the VLM tends to be conservative, repeatedly producing a narrow set of canonical relation descriptions to maintain accuracy. 
To compensate for this limited lexical variety, an LLM rephrases each validated triplet (subject, relation, target) into diverse linguistic forms, aiming to preserve the subject-target correspondence while increasing query diversity.
For example, the triplet (TV, hanging on, wall) can be rephrased into queries such as ``mounted on'' or ``installed on.''
\section{Experiments}
\paragraph{Implementation details.}
For the automated annotation pipeline described in Sec.~\ref{sec:dataset}, we use Qwen3-VL-8B~\cite{bai2025qwen3} as the VLM and GPT-5.2 for predicate rephrasing.
For text-query encoding, we use Jina Embeddings v3, following RelationField~\cite{koch2025relationfield}. The relation transformer follows the ViT-S configuration~\cite{dosovitskiy2020image}, including the transformer depth, number of attention heads, and hidden dimensionality. Our model outputs dense relation predictions at a spatial resolution of \(512 \times 388\).
We train our model on 900 ScanNet++~\cite{yeshwanth2023scannetpp} scenes annotated by the proposed dataset pipeline. Training is performed for 200 epochs on 4 NVIDIA H200 GPUs with a per-GPU batch size of 7. We use AdamW with a learning rate of \(1 \times 10^{-4}\) and cosine learning-rate decay.
\paragraph{Baselines.}
We compare against representative feature-field methods for open-vocabulary 3D scene understanding. 
LangSplat~\cite{langsplat} and OpenGaussians~\cite{wu2024opengaussian} are CLIP-based methods for open-vocabulary object segmentation; at inference time, we directly use the full relation query as the text prompt, since they do not explicitly model object-predicate structure.
RelationField is the closest baseline, as it supports both object-level and predicate-level querying. 
It first localizes the target object using the object field and then refines the prediction using the relation embedding. 
For a more isolated evaluation of relation modeling, we also evaluate RelationField$\dagger$~\cite{koch2025relationfield}, which removes the target-object retrieval stage and localizes regions using only the relation embedding.
Notably, all feature-field baselines require posed RGB images and per-scene optimization, whereas our feed-forward model directly predicts relation fields from unposed multi-view images.

\paragraph{Benchmarks.}
We build a manually annotated benchmark on ScanNet++~\cite{yeshwanth2023scannetpp}, LERF~\cite{kerr2023lerf}, and Replica~\cite{straub2019replica} to evaluate relational understanding in 3D scenes.
For ScanNet++, we use the 26 samples proposed by RelationField and add 14 newly annotated samples. 
For LERF and Replica, we annotate 31 relations from 3 scenes and 48 relations from 8 scenes, respectively. 
Frames are sampled with a moderate stride to avoid overly similar views and ensure sufficient viewpoint diversity. 
The resulting benchmark covers 28 unique predicates.
 We report mIoU as the primary metric and additionally report mAcc, which measures the fraction of queries whose IoU exceeds a threshold of 0.25, providing a complementary view of per-query success rate.
\vspace{-10pt}
\begin{figure*}[t]
    \centering
    \includegraphics[width=\linewidth]{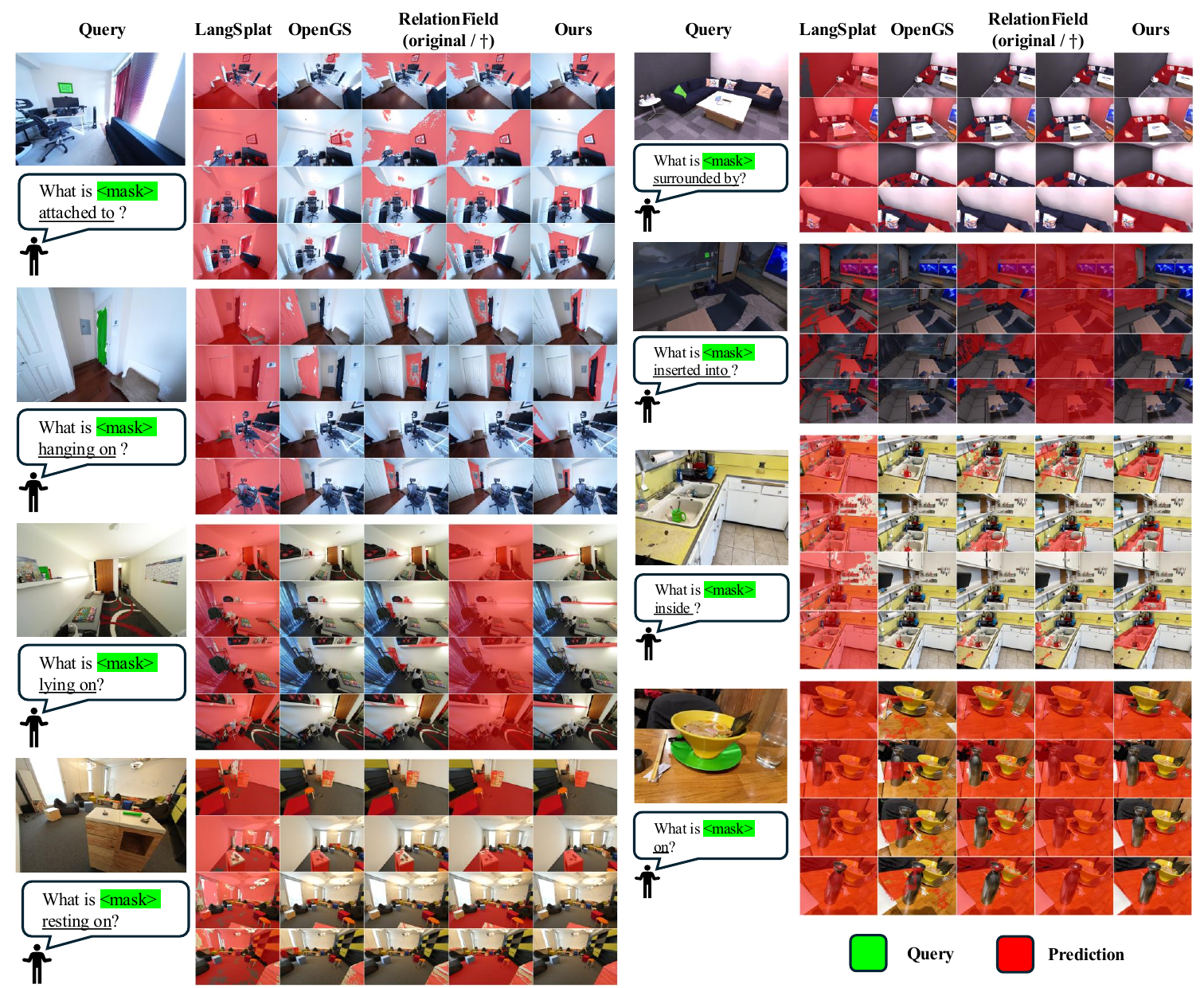}
    \vspace{-10pt}
    \caption{
    Qualitative results on the ScanNet++, Replica, LERF benchmarks. 
    Please zoom for the detail. 
    Predicted masks are overlaid in red on the target views.}
   \label{fig:quali_union}
   \vspace{-15pt}
\end{figure*}

\subsection{Quantitative results}
\vspace{3pt}
\begin{table*}[t]
\centering
\small
\setlength{\tabcolsep}{4pt}
\renewcommand{\arraystretch}{1.05}

\begin{tabular*}{\textwidth}{@{\extracolsep{\fill}}lcccccccc@{}}
\toprule
\multirow{2}{*}{\textbf{Method}}
& \multicolumn{2}{c}{\textbf{Replica}}
& \multicolumn{2}{c}{\textbf{LERF}}
& \multicolumn{2}{c}{\textbf{ScanNet++}}
& \multicolumn{2}{c}{\textbf{Avg.}} \\
\cmidrule(lr){2-3}
\cmidrule(lr){4-5}
\cmidrule(lr){6-7}
\cmidrule(lr){8-9}

& \mIoU & \mAcc
& \mIoU & \mAcc
& \mIoU & \mAcc
& \mIoU & \mAcc \\
\midrule

LangSplat~\cite{langsplat}
& 0.191 & 0.256
& 0.159 & 0.254
& 0.074 & 0.064
& 0.141 & 0.191 \\

OpenGaussian~\cite{wu2024opengaussian}
& \cellcolor{orange!40}0.349 & \cellcolor{yellow!40}0.470
& \cellcolor{orange!40}0.256 & \cellcolor{orange!40}0.378
& 0.162 & 0.254
& \cellcolor{yellow!40}0.255 & \cellcolor{yellow!40}0.367 \\

RelationField~\cite{koch2025relationfield}
& \cellcolor{yellow!40}0.326 & \cellcolor{orange!40}0.537
& \cellcolor{red!40}0.276 & \cellcolor{red!40}0.410
& \cellcolor{orange!40}0.307 & \cellcolor{orange!40}0.491
& \cellcolor{orange!40}0.303 & \cellcolor{orange!40}0.480 \\

RelationField$^\dagger$
& 0.219 & 0.336
& 0.232 & \cellcolor{yellow!40}0.346
& \cellcolor{yellow!40}0.240 & \cellcolor{yellow!40}0.395
& 0.230 & 0.359 \\

\midrule

\textbf{Ours}
& \cellcolor{red!40}0.446 & \cellcolor{red!40}0.640
& \cellcolor{yellow!40}0.254 & 0.332
& \cellcolor{red!40}0.445 & \cellcolor{red!40}0.675
& \cellcolor{red!40}0.382 & \cellcolor{red!40}0.549 \\

\bottomrule
\end{tabular*}
\vspace{-6pt}
\caption{
Quantitative comparison on the Replica, LERF, and ScanNet++ benchmarks.
RelationField$^\dagger$ denotes the relation-only variant without target-object retrieval;
Avg. is the macro average across the three benchmarks.
}
\label{tab:main_comparison}
\vspace{-10pt}
\end{table*}
Table.~\ref{tab:main_comparison} summarizes the segmentation performance across three benchmark splits. 
Our method achieves the highest average mIoU of 0.382 and mAcc of 0.549, outperforming the strongest baseline, RelationField, by substantial margins in both mIoU and mAcc.
Notably, our method does not rely on target object queries at test time and instead leverages relational reasoning alone, which is a key distinction from prior approaches.

On the Replica-based and ScanNet++-based splits, where camera trajectories broadly cover entire indoor scenes rather than focusing on individual objects, open-vocabulary segmentation becomes particularly challenging. 
Our method consistently outperforms all baselines on both splits, demonstrating its robustness in handling diverse viewpoints and cluttered environments.

On the LERF-based split, all methods exhibit lower overall performance due to the densely cluttered tabletop scenes, where numerous small objects are tightly co-located and relational reasoning among them becomes inherently more difficult. 
Even under this challenging setting, our method remains comparable to RelationField, which has access to ground-truth category names at test time.
Furthermore, our method outperforms RelationField$^\dagger$, which relies solely on relation embeddings without object retrieval, confirming that our relational modeling captures meaningful inter-object structure even in cluttered scenarios.

\paragraph{Comparison with video foundation models.}

\begin{table}[t]
\centering
\vspace{-2pt}
\small
\setlength{\tabcolsep}{4pt}
\renewcommand{\arraystretch}{1.05}

\begin{tabularx}{\linewidth}{
    @{}
    l
    *{6}{>{\centering\arraybackslash}X}
    @{}
}
\toprule
\multirow{2}{*}{\textbf{Method}}
& \multicolumn{2}{c}{\textbf{Small} (40)}
& \multicolumn{2}{c}{\textbf{Medium} (39)}
& \multicolumn{2}{c}{\textbf{Large} (40)} \\
\cmidrule(lr){2-3}
\cmidrule(lr){4-5}
\cmidrule(lr){6-7}

& \mIoU & \mAcc
& \mIoU & \mAcc
& \mIoU & \mAcc \\
\midrule

VideoLISA~\cite{video-lisa}
& 0.137 & 0.171
& 0.105 & 0.143
& 0.038 & 0.064 \\

SA2VA~\cite{yuan2025sa2va}
& \cellcolor{yellow!40}0.267
& \cellcolor{yellow!40}0.321
& \cellcolor{yellow!40}0.225
& \cellcolor{yellow!40}0.289
& \cellcolor{yellow!40}0.192
& \cellcolor{yellow!40}0.225 \\

UniPixel~\cite{liu2025unipixel}
& \cellcolor{red!40}0.423
& \cellcolor{orange!40}0.496
& \cellcolor{orange!40}0.328
& \cellcolor{orange!40}0.451
& \cellcolor{orange!40}0.354
& \cellcolor{orange!40}0.475 \\

\midrule

\textbf{Ours}
& \cellcolor{orange!40}0.417
& \cellcolor{red!40}0.550
& \cellcolor{red!40}0.404
& \cellcolor{red!40}0.640
& \cellcolor{red!40}0.366
& \cellcolor{red!40}0.575 \\

\bottomrule
\end{tabularx}
\vspace{2pt}
\caption{
Comparison with video foundation models across camera-baseline ranges.
Small, Medium, and Large denote near-equal groups of the 119 queries ordered by camera baseline.
All methods are given the subject region and relation query without the target category; video baselines process the multi-view images as a video sequence.
}
\label{tab:vs_videollm_results}
\vspace{-5pt}
\end{table}
\vspace{-10pt}
\begin{figure*}[t]
    \centering
    \includegraphics[width=\linewidth]{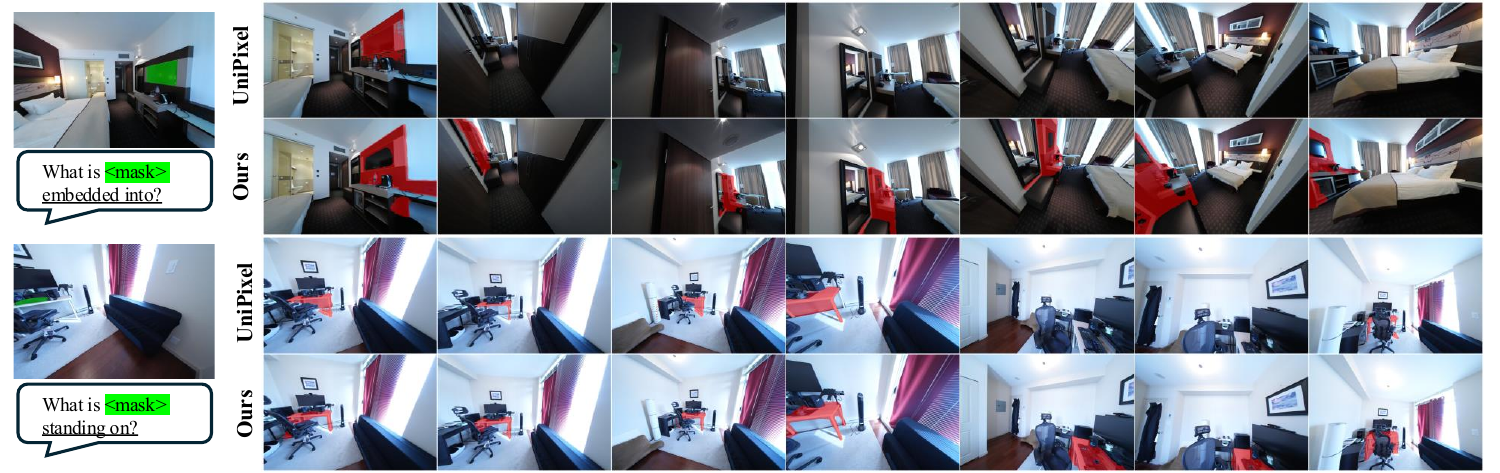}
    \vspace{-10pt}
    \caption{
    Qualitative comparison between our method and UniPixel. 
    Our method produces view-consistent masks, whereas UniPixel fails to maintain consistency across views.
    Please zoom for the detail. 
    Predicted masks are overlaid in red on the target views.
    }
   \label{fig:quali_vs_videoLLM}
   \vspace{-10pt}
\end{figure*}
As shown in Table.~\ref{tab:vs_videollm_results}, we additionally compare against video foundation models by providing the subject region and relation query as input (omitting the target category name) and treating multi-view frames as video input. 
We sort all benchmark samples by camera baseline and divide them into three equal-sized groups. 
Among the compared methods, UniPixel achieves competitive mIoU on the small-baseline setting, but performance generally degrades as the baseline widens, since larger viewpoint changes break the temporal continuity assumptions underlying video representations. 
In contrast, our method maintains stable performance across all baseline ranges. 
We attribute this to the 3D foundation model backbone, which provides multi-view-aware features encoding scene structure rather than relying on frame-to-frame correspondence. 
The result is visualized in Fig.~\ref{fig:quali_vs_videoLLM}.

\subsection{Qualitative results}
Fig.~\ref{fig:quali_union} presents qualitative comparisons against baselines.
LangSplat and OpenGaussian produce fragmented masks with significant spillover into background regions, as CLIP-based features alone lack the granularity to localize relation-relevant areas. RelationField yields more focused predictions but still exhibits noisy boundaries, while RelationField$^\dagger$ tends to activate broadly across the scene due to the absence of target-object retrieval. In contrast, our method produces notably cleaner mask boundaries that tightly conform to the target region.

Moreover, despite not employing any explicit 3D representation such as radiance fields or 3D Gaussian splatting, our predictions remain visually consistent across viewpoints with substantial changes in camera pose. 
We attribute this to the 3D foundation model backbone, which enables the relation transformer to implicitly capture view-invariant relational semantics from multi-view input without explicit 3D reconstruction.

\subsection{Ablations}
\paragraph{Variants of input features.} As an alternative source of geometric cues, we consider using 3D pointmaps, which explicitly represent scene geometry as per-pixel 3D coordinates.
Following~\cite{mv-sam}, we apply standardization to account for scene scale variations, and then use positional encoding to project the coordinates into a higher-dimensional space.
We observe that employing 3D pointmaps leads to a performance drop.
We hypothesize that this is because raw pointmap coordinates, while geometrically informative, lack the semantic cues already embedded in learned features.
Removing DINOv2~\cite{oquab2023dinov2} features leads to a notable drop, as the model loses access to the rich object-level semantics that DINOv2 provides.
The degradation is even more pronounced without Pi3 features, since DINOv2 alone lacks multi-view awareness and cannot capture cross-view geometric correspondence.

\begin{wraptable}{r}{0.51\textwidth}
\vspace{-8pt}
\centering
\small
\setlength{\tabcolsep}{5pt}
\renewcommand{\arraystretch}{1}
\begin{tabular}{lcc}
\toprule
\textbf{Design} & \textbf{mIoU$\uparrow$} & \textbf{mAcc$\uparrow$} \\
\midrule
\textbf{Ours} & \textbf{0.445} & \textbf{0.675} \\
\midrule
\multicolumn{3}{l}{\textit{Input feature variants}}\\
\quad P.E(pointmap) $+$ DINOv2 & 0.370 & 0.568 \\
\quad pi3 only & 0.422 & 0.579 \\
\quad DINOv2 only & 0.370 & 0.579 \\
\midrule
\multicolumn{3}{l}{\textit{Architecture variants}}\\
\quad cross-attn only & 0.399 & 0.579 \\
\quad alternating-attention~\cite{wang2025vggt} & 0.394 & 0.571 \\
\midrule
\multicolumn{3}{l}{\textit{Training objectives}} \\
\quad single-label sigmoid focal loss & 0.406 & 0.604 \\
\midrule
\multicolumn{3}{l}{\textit{Annotation pipeline}} \\
\quad w/o geometric edge filtering & 0.398 & 0.625 \\
\bottomrule
\end{tabular}
\caption{Ablation study on model design and the annotation pipeline, evaluated on ScanNet++ benchmark (40 queries) with similarity threshold $\tau = 0.3$.}
\label{tab:ablation_relation_decoder_design}
\vspace{-10pt}
\end{wraptable}

\paragraph{Variants of relation decoder architecture.}
To validate the design of the relation transformer, we compare against two alternative attention block configurations: a cross-attention-only variant and the alternating attention block proposed by~\cite{wang2025vggt}, which interleaves self-attention and per-frame attention instead of our design. 
As shown in Table~\ref{tab:ablation_relation_decoder_design}, both variants lead to lower performance, demonstrating the effectiveness of our attention block design.

\paragraph{Training objectives.}
Replacing our multi-query sigmoid focal loss with a single-label sigmoid focal loss leads to a notable performance drop, confirming that spatial relations are inherently many-to-one. 
By allowing multiple valid predicates to supervise each target mask simultaneously, the model learns a more robust relational representation and avoids being penalized for semantically plausible alternatives.

\paragraph{Effect of dataset pipeline.}
As shown in Table~\ref{tab:ablation_relation_decoder_design}, removing geometry-based edge filtering causes a performance drop, as the model receives physically implausible supervision such as contact relations between distant objects. 
Additional dataset statistics and analysis are provided in the appendix.

\begin{table}[t]
    \centering
    \small
    \setlength{\tabcolsep}{5pt}
    \begin{tabular}{lccc}
        \toprule
        Subject mask & Mask IoU $\uparrow$ & mIoU $\uparrow$ & mAcc $\uparrow$ \\
        \midrule
        GT mask                  & 1.000              & 0.382          & 0.549          \\
        SAM2, one click          & 0.871  & \textbf{0.383} & \textbf{0.551} \\
        SAM2, one box            & 0.923              & 0.382          & 0.549          \\
        Qwen3-VL + SAM2          & 0.719              & 0.357          & 0.512          \\
        GroundingDINO + SAM2     & 0.587              & 0.322          & 0.454          \\
        \bottomrule
    \end{tabular}
    \vspace{3pt}
    \caption{
    Robustness to subject-mask specification across three benchmarks (119 queries).
    All variants use the same RelationVGGT checkpoint without retraining.
    }
\label{tab:automatic_mask_estimation}
\end{table}
\paragraph{Robustness to subject-mask specification.}
We replace the ground-truth subject mask with promptable or automatically estimated masks without retraining the model.
As shown in Table~\ref{tab:automatic_mask_estimation}, single-click or box prompting with SAM2 preserves essentially the same performance as the ground-truth mask despite imperfect mask overlap, indicating that precise subject boundaries are not required when the intended instance is correctly specified.
We further evaluate fully automatic subject grounding using Qwen3-VL-8B+SAM2 and GroundingDINO+SAM2.
The former retains most of the ground-truth-mask performance without manual annotation, whereas the latter shows a larger degradation.
We attribute this primarily to subject-instance ambiguity rather than boundary precision, as 58\% of reference views contain multiple instances of the queried category.

\paragraph{Robustness to predicate expressions.}
\begin{table}[t]
    \centering
    \small
    \setlength{\tabcolsep}{6pt}
    \renewcommand{\arraystretch}{1.05}
    \begin{tabular}{lccc}
        \toprule
        Predicate expression
        & $n$
        & $\Delta$mIoU $\uparrow$
        & $\Delta$mAcc@0.25 $\uparrow$ \\
        \midrule
        Meaning-preserving synonyms (set A) & 119 & -0.006 & -0.006 \\
        Meaning-preserving synonyms (set B) & 119 & -0.004 & -0.001 \\
        Rare synonym expressions            & 113 & -0.022 & -0.042 \\
        Unseen synonym expressions          & 119 & -0.031 & -0.040 \\
        \bottomrule
    \end{tabular}
    \vspace{3pt}
    \caption{
    Robustness to predicate expressions.
    We report paired per-query changes relative to the original predicate queries.
    The two synonym sets use independently constructed meaning-preserving alternatives.
    Rare expressions use the least frequent suitable synonym observed in training (stem frequency $\leq 20$ for 94/113 queries; six queries without a suitable alternative are excluded), whereas unseen expressions use synonyms whose stems never occur during training.
    }
    \label{tab:predicate_robustness}
\end{table}
We evaluate lexical robustness by replacing the benchmark predicates with meaning-preserving alternatives while keeping the scene, subject mask, and target fixed.
As shown in Table~\ref{tab:predicate_robustness}, the model remains robust to paraphrased predicate expressions, with only minor degradation for common synonyms and limited degradation even for rare or previously unseen expressions.
These results indicate that relation prediction is not tied to exact predicate strings observed during training, but generalizes across semantically equivalent formulations.
\section{Conclusion}
In this work, we formulated \emph{3D spatial relation segmentation} in a feed-forward, pose-free multi-view setting, where the goal is to localize target objects that satisfy a spatial relation with respect to a visually specified subject using the underlying 3D scene structure.
To tackle this, we proposed RelationVGGT, a feed-forward framework that integrates 3D geometry priors with 2D visual foundation models for subject-conditioned relational segmentation across views, without per-scene optimization or known camera poses. We also presented a fully automated data generation pipeline built on ScanNet++ with VLM and LLM integration, enabling scalable multi-view consistent annotations.

\paragraph{Limitations and future work.}
First, our data generation pipeline currently depends on datasets with instance-level annotations such as ScanNet++; extending it to unannotated or in-the-wild scenes is an important direction.
Second, while we adopt Set-of-Mark prompting to extract relations via VLMs, exploring more accurate and efficient annotation strategies remains worthwhile. 
Third, our framework models object-object spatial relations, but could be extended to functional relations or combined with panoptic segmentation branches to jointly predict scene graph nodes and edges.

% \begin{ack}
% Use unnumbered first level headings for the acknowledgments. All acknowledgments

% Do {\bf not} include this section in the anonymized submission, only in the final paper. You can use the \texttt{ack} environment provided in the style file to automatically hide this section in the anonymized submission.
% \end{ack}

\small
\bibliographystyle{unsrt}  % order of appearanc
\bibliography{references}

%%%%%%%%%%%%%%%%%%%%%%%%%%%%%%%%%%%%%%%%%%%%%%%%%%%%%%%%%%%%

\appendix
% Technical appendices with additional results, figures, graphs, and proofs may be submitted with the paper submission before the full submission deadline (see above). You can upload a ZIP file for videos or code, but do not upload a separate PDF file for the appendix. There is no page limit for the technical appendices. 

% Note: Think of the appendix as ``optional reading'' for reviewers. The paper must be able to stand alone without the appendix; for example, adding critical experiments that support the main claims to an appendix is inappropriate. 

%%%%%%%%%%%%%%%%%%%%%%%%%%%%%%%%%%%%%%%%%%%%%%%%%%%%%%%%%%%%

\newpage
% \section{Technical appendices and supplementary material}
% RelationVGGT appendix 
\section{Dataset Construction and Quality}
\label{app:dataset_quality}

\subsection{Relation distribution}
\label{app:relation_distribution}

\begin{figure}[H]
    \centering
    \includegraphics[width=\linewidth]{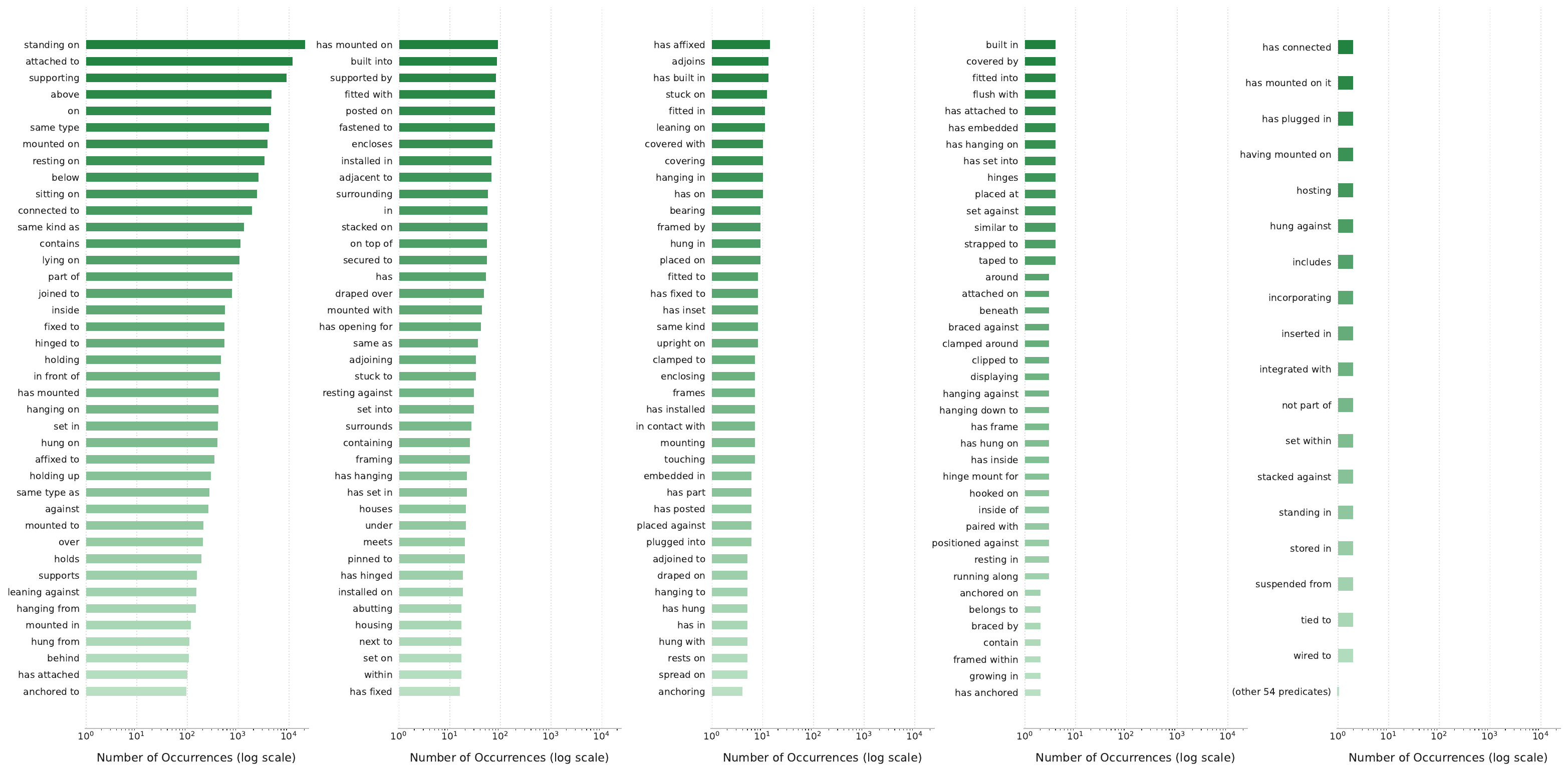}
    \caption{
        Distribution of the final relation predicates used for training.
        The horizontal axis shows the number of occurrences on a logarithmic scale; 54 additional
        low-frequency predicates are grouped for visualization.
    }
    \label{fig:relation_distribution}
\end{figure}

Figure~\ref{fig:relation_distribution} shows the final predicate distribution in our training
annotations. The distribution is strongly long-tailed: frequent relations such as
\emph{standing on}, \emph{attached to}, and \emph{supporting} coexist with a large number of
low-frequency expressions, reflecting the lexical diversity of the open-vocabulary annotations and
motivating our multi-query supervision rather than a fixed predicate classifier.

\subsection{Human audit of annotation quality}
\label{app:annotation_audit}

To quantify the quality of the automatically generated supervision, two authors independently
audited all 4,695 directed relation proposals produced across 25 ScanNet++ scenes. For each proposal,
an annotator was shown the relation proposition together with its supporting views, but not the
pipeline's keep/remove decision, and assigned one of three labels: \emph{correct}, \emph{incorrect},
or \emph{ambiguous}. We exclude ambiguous judgments from the denominators of the quality rates
reported below. Since the audit covers a finite set of scenes, we compute 95\% confidence intervals
with a scene-level cluster bootstrap using 20,000 resamples. Unless otherwise noted, each reported
rate is the mean of the two annotators.

\begin{table}[H]
    \centering
    \begin{tabular}{lrr}
        \toprule
        Stage & Relations after stage & \% of Stage 1 \\
        \midrule
        Stage 1: initial aggregation
        & 4,695 & 100.0 \\
        Stage 2: filtering
        & 770 & 16.4 \\
        \bottomrule
    \end{tabular}
    \vspace{3pt}
    \caption{
        Number of relation proposals that survive the annotation filtering pipeline. (25 scenes as samples)
        Stage 2 combines multi-view consistency with the geometry-based filtering described in
        Sec.~\ref{sec:dataset}.
    }
    \label{tab:annotation_stage_counts}
\end{table}
\vspace{-10pt}

\subsubsection{How many proposals survive filtering?}
\label{app:annotation_survival}

Table~\ref{tab:annotation_stage_counts} summarizes the number of relation proposals before and after
the multi-view and geometric filtering stage.
Stage 2 removes 3,925 of the 4,695 aggregated proposals. This filtering is intentionally conservative:
a relation supported by fewer than three frames is discarded even when it may be correct, because a
false relation that survives the filter becomes direct training supervision, whereas a removed correct
relation primarily reduces coverage.

\subsubsection{Human judgment versus filtering decision}
\label{app:annotation_confusion}

Table~\ref{tab:annotation_human_filter} compares the human judgments with the Stage-2 keep/remove
decision. Counts are reported as annotator 1 / annotator 2. The total numbers differ because ambiguous
judgments are excluded from the table and all derived rates; the two annotators marked 33 and 409
proposals as ambiguous, respectively.

The filter removes the large majority of proposals judged incorrect, although it also discards a
substantial number of correct proposals. The latter is expected from the conservative $K=3$
multi-view agreement requirement, which rejects relations that are visible or confidently recognized
in only one or two frames.

\begin{table}[t]
    \centering
    \caption{
        Human judgment versus the Stage-2 filtering decision.
        Counts are reported as annotator 1 / annotator 2; ambiguous judgments are excluded.
    }
    \label{tab:annotation_human_filter}
    \begin{tabular}{lrrr}
        \toprule
        Human judgment
        & Kept by Stage 2
        & Removed by Stage 2
        & Total \\
        \midrule
        Correct
        & 506 / 421
        & 1,151 / 1,173
        & 1,657 / 1,594 \\
        Incorrect
        & 257 / 198
        & 2,748 / 2,494
        & 3,005 / 2,692 \\
        \midrule
        Total
        & 763 / 619
        & 3,899 / 3,667
        & 4,662 / 4,286 \\
        \bottomrule
    \end{tabular}
\end{table}

\subsubsection{Annotation quality}
\label{app:annotation_quality}

We summarize annotation quality in Table~\ref{tab:annotation_quality}. Precision after Stage 1 is the
fraction of aggregated proposals judged correct, while precision after Stage 2 is the fraction of
retained training relations judged correct. ``Incorrect proposals removed'' measures the fraction of
human-judged incorrect proposals discarded by Stage 2, and ``Correct proposals retained'' measures
the fraction of human-judged correct proposals that survive the filtering stage.

\begin{table}[t]
    \centering
    \caption{
        Human evaluation of the annotation pipeline on 25 ScanNet++ scenes.
        Confidence intervals are obtained with scene-level cluster bootstrap using 20,000 resamples.
    }
    \vspace{3pt}
    \label{tab:annotation_quality}
    \begin{tabular}{lccc}
        \toprule
        Statistic
        & Mean (\%)
        & 95\% CI
        & Annotator 1 / 2 (\%) \\
        \midrule
        Precision after Stage 1
        & 36.4
        & [32.2, 39.2]
        & 35.5 / 37.2 \\
        Precision after Stage 2
        & \textbf{67.2}
        & \textbf{[60.3, 73.5]}
        & 66.3 / 68.0 \\
        Incorrect proposals removed
        & \textbf{92.0}
        & \textbf{[89.1, 94.0]}
        & 91.4 / 92.6 \\
        Correct proposals retained
        & 28.5
        & [22.8, 37.2]
        & 30.5 / 26.4 \\
        \bottomrule
    \end{tabular}
\end{table}

Filtering increases proposal precision from 36.4\% to 67.2\% and removes 92.0\% of proposals judged
incorrect. The 28.5\% correct-retention rate should not be interpreted as recall over all true scene
relations; it only measures retention within the Stage-1 proposal pool.

The two annotators agree on 83.1\% of all proposals (Cohen's $\kappa=0.670$), with a reported
$\kappa=0.816$ when ambiguous judgments are excluded. These results indicate that the filtering
pipeline substantially improves the precision of the automatically generated supervision, while also
showing that the retained set remains noisy and intentionally sacrifices coverage for reliability.

\section{Implementation Details}
\label{app:implementation_details}

\subsection{Training details}
\label{app:training_details}

For each training sample, let $\mathcal{Q}$ denote the set of compatible phrasings generated from
the annotated spatial relation. All queries $q \in \mathcal{Q}$ describe the same relation and are
therefore supervised using the same ground-truth target mask $M_t$. Rather than treating these
phrasings as mutually exclusive classes, we evaluate each query independently as a binary relation
prediction.

For a target-view pixel $u$ and query $q$, we denote the predicted probability by
\begin{equation}
    p = S_t(u;q),
\end{equation}
and the corresponding ground-truth label by
\begin{equation}
    y = M_t(u) \in \{0,1\}.
\end{equation}
The focal loss for each pixel--query pair is
\begin{equation}
    \ell_{\mathrm{focal}}(p,y)
    =
    -\alpha_y (1-p_y)^\gamma \log p_y,
\end{equation}
where
\begin{equation}
    p_y =
    \begin{cases}
        p, & y=1,\\
        1-p, & y=0,
    \end{cases}
    \qquad
    \alpha_y =
    \begin{cases}
        \alpha, & y=1,\\
        1-\alpha, & y=0.
    \end{cases}
\end{equation}
We use $\alpha=0.75$ and $\gamma=1.5$ in all experiments.

Let $\Omega_t$ denote the set of pixels in the target view. The multi-query focal loss is obtained
by averaging over both pixels and compatible queries:
\begin{equation}
    \mathcal{L}_{\mathrm{focal}}^{\mathrm{multi\text{-}q}}
    =
    \frac{1}{|\mathcal{Q}|\,|\Omega_t|}
    \sum_{q\in\mathcal{Q}}
    \sum_{u\in\Omega_t}
    \ell_{\mathrm{focal}}
    \left(S_t(u;q), M_t(u)\right).
\end{equation}
Similarly, the Dice loss is computed independently for each compatible query against the same
target mask and averaged over $\mathcal{Q}$:
\begin{equation}
    \mathcal{L}_{\mathrm{dice}}
    =
    \frac{1}{|\mathcal{Q}|}
    \sum_{q\in\mathcal{Q}}
    \mathcal{L}_{\mathrm{dice}}
    \left(S_t(\cdot;q), M_t\right).
\end{equation}
The overall mask supervision is
\begin{equation}
    \mathcal{L}_{\mathrm{mask}}
    =
    w_{\mathrm{focal}}
    \mathcal{L}_{\mathrm{focal}}^{\mathrm{multi\text{-}q}}
    +
    w_{\mathrm{dice}}
    \mathcal{L}_{\mathrm{dice}}.
\end{equation}
This formulation allows multiple semantically valid descriptions of the same spatial relation to
serve as positive supervision without forcing competition between them. Further implementation
details are available in our released code.

\subsection{Inference efficiency}
\label{app:inference_efficiency}

Table~\ref{tab:inference_efficiency} compares inference efficiency with RelationField, which requires scene-specific optimization before relation queries can be evaluated.
On an NVIDIA L40S, our method requires
26\,ms for inference in the timing setup used in our evaluation, whereas RelationField requires 288\,s for a relation query and more than 3\,hours of per-scene optimization.

The timing comparison reflects the different operating regimes of the two methods: RelationField constructs a scene-specific representation, while our model performs feed-forward inference on unseen scenes without scene-level optimization.

\begin{table}[H]
    \centering
    \begin{tabular}{lcc}
        \toprule
        Method
        & Query inference
        & Per-scene optimization \\
        \midrule
        RelationField
        & 288\,s
        & $>$3\,h \\
        Ours
        & 26\,ms
        & None \\
        \bottomrule
    \end{tabular}
    \caption{
        Inference-time comparison with RelationField on an NVIDIA L40S.
        RelationField additionally requires scene-specific optimization before query-time inference.
    }
    \label{tab:inference_efficiency}
\end{table}

\subsection{Annotation prompts}
\label{app:annotation_prompts}

Figures~\ref{fig:prompt-extraction} and~\ref{fig:prompt-relation-refinement} show the exact prompts
used in our annotation pipeline. For visual relation extraction, the VLM receives a single
SoM-tagged image and is instructed to identify only visible tagged instances and return directed,
open-vocabulary relations in a structured JSON format. The prompt explicitly discourages unsupported
or ambiguous relations, preserves the subject$\rightarrow$object direction, and provides relation
families only as references rather than as a closed predicate vocabulary.

The subsequent relation-refinement stage is deliberately restricted to text-only, category-level
information. Given the subject category, input relation, object category, and relation family, the
LLM is asked to produce a short, natural predicate phrase while preserving both the original semantic
intent and the relation family. It receives no image or scene geometry and is explicitly prohibited
from inferring instance-level pose or layout, which separates lexical refinement from the visual
evidence used to establish the relation itself.

% The following source contains the two exact prompt figures.
% Remove the final \clearpage in LLM_prompt_appendix.tex if it creates an unwanted page break.
\definecolor{promptbg}{HTML}{EDEDED}
\definecolor{promptframe}{HTML}{888888}
\definecolor{rolekey}{HTML}{185FA5}
\definecolor{strval}{HTML}{2D6A1F}

\setlength{\textfloatsep}{8pt}
\setlength{\floatsep}{8pt}
\setlength{\intextsep}{8pt}

\begin{figure}[!t]
\centering

\begin{tcolorbox}[
    colback=promptbg,
    colframe=promptframe,
    boxrule=0.5pt,
    arc=3pt,
    left=8pt, right=8pt, top=6pt, bottom=6pt,
    fontupper=\footnotesize,
    width=0.96\textwidth,
]
{\color{rolekey}\textbf{[System]}} You are given one SoM-tagged indoor scene image. Identify visible tagged objects and infer visually grounded open-vocabulary relationships between them.

\vspace{4pt}
\textbf{Input.} One indoor scene image in which each visible object instance has exactly one integer tag placed near its segmented region.

\vspace{4pt}
\textbf{Output.} Return strict JSON only with two fields: \texttt{objects} and \texttt{relationships}.  
\texttt{objects} maps each visible \texttt{tag\_id} to a short singular object name.  
\texttt{relationships} is a list of \texttt{\{s\_id, o\_id, predicates\}} entries.

\vspace{4pt}
\textbf{Object rules.} Use only \texttt{visible tagged instances}. Ignore untagged, hidden, guessed, or heavily ambiguous objects. The SoM image is the primary evidence source; tags are grounding aids for object identity.

\vspace{4pt}
\textbf{Relationship rules.} Infer only visually supported relations. Do not force a relation for every pair. Merge valid non-redundant predicates for the same ordered pair. Predicates are open-vocabulary, and the listed examples are only references, not a fixed label set.

\vspace{4pt}
\textbf{Core constraints.} Avoid vague predicates such as \texttt{related\_to} or \texttt{associated\_with}. Do not use \texttt{left}, \texttt{right}, \texttt{in\_front\_of}, or \texttt{behind}. Preserve subject$\rightarrow$object direction. If evidence is weak, blurry, occluded, or ambiguous, omit the relation.

\vspace{4pt}
\textbf{Relation families.}
\begin{description}[
    leftmargin=4.15cm,
    labelwidth=3.85cm,
    labelsep=0.3cm,
    itemsep=1pt,
    topsep=2pt,
    parsep=0pt,
    partopsep=0pt,
    font=\normalfont\bfseries
]
    \item[1. composition] \texttt{part\_of}, \texttt{component\_of}
    \item[2. topology] \texttt{inside}, \texttt{contains}, \texttt{attached\_to}
    \item[3. support-contact] \texttt{on}, \texttt{resting\_on}, \texttt{mounted\_on}
    \item[4. vertical directional] \texttt{above}, \texttt{below}
    \item[5. proximity/adjacency] \texttt{near}, \texttt{adjacent\_to}, \texttt{next\_to}
    \item[6. comparative/similarity] \texttt{same\_type\_as}, \texttt{same\_color\_as}, \texttt{larger\_than}
\end{description}

\vspace{4pt}
\textbf{Example output.}

\vspace{2pt}
\begin{tcolorbox}[
    colback=white,
    colframe=promptframe,
    boxrule=0.3pt,
    arc=2pt,
    left=5pt, right=5pt, top=3pt, bottom=3pt,
    fontupper=\ttfamily\scriptsize,
]
\{\\
\hspace*{0.8em}{\color{rolekey}"objects"}: \{ {\color{strval}"4"}: {\color{strval}"floor"}, {\color{strval}"7"}: {\color{strval}"table"} \},\\
\hspace*{0.8em}{\color{rolekey}"relationships"}: [\\
\hspace*{1.6em}\{ {\color{rolekey}"s\_id"}: {\color{strval}"7"}, {\color{rolekey}"o\_id"}: {\color{strval}"4"}, {\color{rolekey}"predicates"}: [{\color{strval}"standing\_on"}] \}\\
\hspace*{0.8em}]\\
\}
\end{tcolorbox}
\end{tcolorbox}

\caption{Prompt for grounded open-vocabulary relationship extraction.}
\label{fig:prompt-extraction}
\end{figure}

% =========================================================
% Figure 2 (single page)
% =========================================================
\begin{figure}[!t]
\centering
\begin{tcolorbox}[
    colback=promptbg,
    colframe=promptframe,
    boxrule=0.5pt,
    arc=3pt,
    left=10pt, right=10pt, top=8pt, bottom=8pt,
    fontupper=\small,
    width=0.96\textwidth,
]
{\color{rolekey}\textbf{[System]}} You are a commonsense relation rephraser for object-category triples.

\medskip
\textbf{Input.} A structured triple with four fields: \texttt{subject\_category}, \texttt{relation}, \texttt{object\_category}, and \texttt{relation\_family}.

\medskip
\textbf{Task.} Refine the relation so that it sounds more natural, idiomatic, and category-appropriate, using only \texttt{category-level commonsense}.

\medskip
\textbf{Scope.} Text-only task. No image, no scene geometry, and no instance-level evidence. Do not infer pose, layout, attachment, or visibility.

\medskip
\textbf{Core objective.} Improve awkward or unidiomatic wording while preserving the original semantic intent. Preserve \texttt{relation\_family} exactly. Make only conservative, same-family refinements, and avoid hallucinated edits.

\medskip
\textbf{Relation-family rule.} The input relation belongs to one fixed family. Improve wording only within that family. Never cross families, even if another family seems plausible.

\medskip
\textbf{Relation families.}
\begin{description}[
    leftmargin=4.35cm,
    labelwidth=3.95cm,
    labelsep=0.4cm,
    itemsep=2pt,
    topsep=4pt,
    font=\normalfont\bfseries
]
    \item[1. composition] part-whole, component, membership.
    \item[2. topology] containment, enclosure, attachment.
    \item[3. support-contact] resting, hanging, leaning, mounting.
    \item[4. vertical directional] vertical ordering only.
    \item[5. proximity/adjacency] local closeness.
    \item[6. comparative/similarity] size, color, shape, material.
\end{description}

\medskip
\textbf{Examples.}

\vspace{2pt}
\begin{tcolorbox}[
    colback=white,
    colframe=promptframe,
    boxrule=0.3pt,
    arc=2pt,
    left=6pt, right=6pt, top=4pt, bottom=4pt,
    fontupper=\ttfamily\footnotesize,
]
(\texttt{TV}, \texttt{hanging\_on}, \texttt{wall}, \textit{support-contact}) $\rightarrow$ {\color{strval}\textbf{mounted\_on}}\\
(\texttt{pillow}, \texttt{standing\_on}, \texttt{bed}, \textit{support-contact}) $\rightarrow$ {\color{strval}\textbf{lying\_on}}\\
(\texttt{chair}, \texttt{located\_at}, \texttt{table}, \textit{proximity/adjacency}) $\rightarrow$ {\color{strval}\textbf{next\_to}}
\end{tcolorbox}

\medskip
\textbf{Lexical constraints.} \texttt{refined\_relation} must be a short predicate-style phrase (e.g., \texttt{mounted\_on}, \texttt{lying\_on}, \texttt{next\_to}, \texttt{similar\_to}). Do not include the subject or object noun. Avoid clauses such as \textit{``attached to the wall''}. Preserve subject$\rightarrow$object direction.

\medskip
\textbf{Output (JSON only).}

\vspace{2pt}
\begin{tcolorbox}[
    colback=white,
    colframe=promptframe,
    boxrule=0.3pt,
    arc=2pt,
    left=6pt, right=6pt, top=4pt, bottom=4pt,
    fontupper=\ttfamily\footnotesize,
]
\{\\
\hspace*{1em}{\color{rolekey}"subject\_category"}: {\color{strval}"<subject\_category>"},\\
\hspace*{1em}{\color{rolekey}"input\_relation"}: {\color{strval}"<original\_relation>"},\\
\hspace*{1em}{\color{rolekey}"refined\_relation"}: {\color{strval}"<same-family naturalized relation>"},\\
\hspace*{1em}{\color{rolekey}"object\_category"}: {\color{strval}"<object\_category>"},\\
\hspace*{1em}{\color{rolekey}"relation\_family"}: {\color{strval}"<same as input>"},\\
\hspace*{1em}{\color{rolekey}"changed"}: <true {\textbar} false>,\\
\hspace*{1em}{\color{rolekey}"confidence"}: {\color{strval}"<high {\textbar} medium {\textbar} low>"}\\
\}
\end{tcolorbox}
\end{tcolorbox}
\caption{Prompt for relation refinement.}
\label{fig:prompt-relation-refinement}
\end{figure}

\clearpage

\section{Additional Experiments}
\label{app:additional_experiments}

We report three analyses that complement the results in the main paper: comparisons with closely
related 3D referring-segmentation methods, robustness when the reference subject is not visible in a
target view, and controls in which the predicate meaning is deliberately changed. Experiments already
reported in the main paper, including subject-mask specification and meaning-preserving predicate
rephrasings, are not repeated here.

\subsection{Comparison with 3D referring-segmentation methods}
\label{app:referring_baselines}

We additionally compare with MVGGT~\cite{mvggt} and ReferSplat~\cite{refersplat}, two closely related 3D referring-segmentation methods. 
Their interfaces differ from ours: neither method accepts a visual subject mask, so the subject must instead be specified through language. 
We therefore adapt each method using its native text-query interface.

\paragraph{MVGGT.}
We evaluate the released checkpoint zero-shot. Each benchmark triplet
$(\text{subject}, \text{relation}, \text{target})$ is converted into a ScanRefer-style referring
expression, following the linguistic form used to train MVGGT. For example,
$(\text{backpack}, \text{leaning against}, \text{wall})$ is expressed as
``this is a wall. it has the backpack leaning against it.'' We evaluate six query templates and report
the best-performing template. In this adaptation, the target category is explicitly named in the text.

\paragraph{ReferSplat.}
ReferSplat requires per-scene optimization, so we fit its Gaussian representation on each benchmark
scene and train its referring branch using the method's native pipeline. At evaluation time, we hide
the target category by replacing it with a generic noun, e.g.,
``the object that the cushion is lying on.'' Thus, the query specifies the subject and relation while
withholding the target category.

\begin{table}[t]
    \centering
    \resizebox{\linewidth}{!}{
    \begin{tabular}{lcccc}
        \toprule
        Method
        & Subject specification
        & Target category in input
        & Test-scene optimization
        & Avg. mIoU / mAcc@0.25 \\
        \midrule
        MVGGT
        & Text
        & Yes
        & No
        & 0.153 / 0.217 \\
        ReferSplat
        & Text
        & No
        & Yes
        & 0.213 / 0.263 \\
        Ours
        & Visual mask
        & No
        & No
        & \textbf{0.382 / 0.549} \\
        \bottomrule
    \end{tabular}}
    \vspace{3pt}
    \caption{
        Comparison with closely related 3D referring-segmentation methods.
        The methods differ in their native input interfaces and optimization protocols; in particular,
        MVGGT receives the target category in the evaluated text template, whereas ReferSplat and
        our method do not.
    }
    \label{tab:referring_baselines}
\end{table}

As shown in Table~\ref{tab:referring_baselines}, our method achieves substantially higher average segmentation accuracy than the adapted referring-segmentation baselines. 
A key difference lies in how the reference subject is specified. 
In text-only referring, the subject must itself be grounded from a noun phrase, which can be ambiguous when multiple instances of the same category are present in the scene. 
Our formulation instead visually specifies the subject, explicitly resolving this instance-level ambiguity and allowing the model to focus on identifying the target from the relation.
This comparison is not a matched-input ablation, since the methods also differ in query form, pretraining data, and test-scene optimization; nevertheless, the results highlight the practical advantage of separating subject grounding from relation-conditioned target localization.

\subsection{Subject visibility across target views}
\label{app:subject_visibility}

The subject is specified only in the reference view and may not be visible in every target view. To
evaluate this case explicitly, we label every scored target view in the ScanNet++ benchmark by
whether the subject is visible and re-score both our method and RelationField on the two subsets.

\begin{table}[t]
    \centering
    \small
    \setlength{\tabcolsep}{5pt}
    \begin{tabular}{@{}lccccc@{}}
        \toprule
        & & \multicolumn{2}{c}{Ours}
        & \multicolumn{2}{c}{RelationField} \\
        \cmidrule(lr){3-4}
        \cmidrule(lr){5-6}
        Target-view subset
        & Views
        & mIoU
        & mAcc
        & mIoU
        & mAcc \\
        \midrule
        Subject visible
        & 257
        & 0.440
        & 0.677
        & 0.318
        & 0.510 \\
        Subject not visible
        & 23
        & 0.507
        & 0.652
        & 0.159
        & 0.217 \\
        \bottomrule
    \end{tabular}
    \vspace{3pt}
    \caption{
        Performance on ScanNet++ target views grouped by subject visibility.
        The subject is always specified in the reference view; visibility here refers only to the
        evaluated target view.
    }
    \label{tab:subject_visibility}
\end{table}

Our method remains effective when the subject is absent from the evaluated target view. This behavior
is consistent with the model design: the subject is grounded in the reference view, while cross-view
attention allows target-view tokens to access subject-conditioned evidence from the multi-view set.
We do not interpret the numerically higher mIoU on the not-visible subset as an intrinsic advantage,
since this subset contains only 23 views and may differ in target size and difficulty.

\subsection{Sensitivity to meaning-changing predicate controls}
\label{app:meaning_changing_controls}

The main paper evaluates meaning-preserving predicate rephrasings. Here, we provide a complementary control in which the predicate meaning is deliberately changed while the scene, subject mask, and original target annotation are held fixed. 
This tests whether the prediction is actually sensitive to the relational text rather than simply reproducing a target independent of the query.
The results are summarized in Table~\ref{tab:meaning_changing_controls}.

\begin{table}[t]
    \centering
    \caption{
        Meaning-changing predicate controls.
        $\Delta$mIoU is measured relative to the original benchmark predicate while keeping the
        original target mask fixed. These controls therefore measure query sensitivity rather than
        segmentation accuracy for the substituted relation itself.
    }
    \label{tab:meaning_changing_controls}
    \begin{tabular}{lcc}
        \toprule
        Predicate substitution
        & $n$
        & $\Delta$mIoU \\
        \midrule
        Random relation from an incompatible family
        & 119
        & $-0.158$ {\small $[-0.207,-0.111]$} \\
        Relation that changes the correct target
        & 118
        & $-0.276$ \\
        \bottomrule
    \end{tabular}
\end{table}

Replacing the predicate with a random relation from an incompatible family reduces mIoU by 0.158.
When the substituted relation is selected so that the correct target changes, the degradation increases
to 0.276. In contrast to the meaning-preserving substitutions reported in the main paper, these
meaning-changing controls substantially alter the prediction. This supports that the model responds
to the semantic content of the relation query rather than only to the visual subject or a fixed
relation label.

\newpage
\section*{NeurIPS Paper Checklist}

%%% BEGIN INSTRUCTIONS %%%
The checklist is designed to encourage best practices for responsible machine learning research, addressing issues of reproducibility, transparency, research ethics, and societal impact. Do not remove the checklist: {\bf The papers not including the checklist will be desk rejected.} The checklist should follow the references and follow the (optional) supplemental material.  The checklist does NOT count towards the page
limit. 

Please read the checklist guidelines carefully for information on how to answer these questions. For each question in the checklist:
\begin{itemize}
    \item You should answer \answerYes{}, \answerNo{}, or \answerNA{}.
    \item \answerNA{} means either that the question is Not Applicable for that particular paper or the relevant information is Not Available.
    \item Please provide a short (1--2 sentence) justification right after your answer (even for \answerNA). 
   % \item {\bf The papers not including the checklist will be desk rejected.}
\end{itemize}

{\bf The checklist answers are an integral part of your paper submission.} They are visible to the reviewers, area chairs, senior area chairs, and ethics reviewers. You will also be asked to include it (after eventual revisions) with the final version of your paper, and its final version will be published with the paper.

The reviewers of your paper will be asked to use the checklist as one of the factors in their evaluation. While \answerYes{} is generally preferable to \answerNo{}, it is perfectly acceptable to answer \answerNo{} provided a proper justification is given (e.g., error bars are not reported because it would be too computationally expensive'' or ``we were unable to find the license for the dataset we used''). In general, answering \answerNo{} or \answerNA{} is not grounds for rejection. While the questions are phrased in a binary way, we acknowledge that the true answer is often more nuanced, so please just use your best judgment and write a justification to elaborate. All supporting evidence can appear either in the main paper or the supplemental material, provided in appendix. If you answer \answerYes{} to a question, in the justification please point to the section(s) where related material for the question can be found.

IMPORTANT, please:
\begin{itemize}
    \item {\bf Delete this instruction block, but keep the section heading ``NeurIPS Paper Checklist"},
    \item  {\bf Keep the checklist subsection headings, questions/answers and guidelines below.}
    \item {\bf Do not modify the questions and only use the provided macros for your answers}.
\end{itemize}

%%% END INSTRUCTIONS %%%

\begin{enumerate}

\item {\bf Claims}
    \item[] Question: Do the main claims made in the abstract and introduction accurately reflect the paper's contributions and scope?
    \item[] Answer: \answerYes{} % Replace by \answerYes{}, \answerNo{}, or \answerNA{}.
    \item[] Justification: We claim the main contributions.

\item {\bf Limitations}
    \item[] Question: Does the paper discuss the limitations of the work performed by the authors?
    \item[] Answer: \answerYes{}{} % Replace by \answerYes{}, \answerNo{}, or \answerNA{}.
    \item[] Justification: We discuss them at conclusion sections.

\item {\bf Theory assumptions and proofs}
    \item[] Question: For each theoretical result, does the paper provide the full set of assumptions and a complete (and correct) proof?
    \item[] Answer: \answerYes{} % Replace by \answerYes{}, \answerNo{}, or \answerNA{}.
    \item[] Justification: We provide proofs.

    \item {\bf Experimental result reproducibility}
    \item[] Question: Does the paper fully disclose all the information needed to reproduce the main experimental results of the paper to the extent that it affects the main claims and/or conclusions of the paper (regardless of whether the code and data are provided or not)?
    \item[] Answer: \answerYes{} % Replace by \answerYes{}, \answerNo{}, or \answerNA{}.
    \item[] Justification: We will provide open-source codes.

\item {\bf Open access to data and code}
    \item[] Question: Does the paper provide open access to the data and code, with sufficient instructions to faithfully reproduce the main experimental results, as described in supplemental material?
    \item[] Answer: \answerNA{} % Replace by \answerYes{}, \answerNo{}, or \answerNA{}.
    \item[] Justification: We will provide it later.

\item {\bf Experimental setting/details}
    \item[] Question: Does the paper specify all the training and test details (e.g., data splits, hyperparameters, how they were chosen, type of optimizer) necessary to understand the results?
    \item[] Answer: \answerYes{}{} % Replace by \answerYes{}, \answerNo{}, or \answerNA{}.
    \item[] Justification: Yes it does at experiments section.

\item {\bf Experiment statistical significance}
    \item[] Question: Does the paper report error bars suitably and correctly defined or other appropriate information about the statistical significance of the experiments?
    \item[] Answer: \answerNA{} % Replace by \answerYes{}, \answerNo{}, or \answerNA{}.
    \item[] Justification: Not related well.

\item {\bf Experiments compute resources}
    \item[] Question: For each experiment, does the paper provide sufficient information on the computer resources (type of compute workers, memory, time of execution) needed to reproduce the experiments?
    \item[] Answer: \answerYes{} % Replace by \answerYes{}, \answerNo{}, or \answerNA{}.
    \item[] Justification: About GPU hours.
    
\item {\bf Code of ethics}
    \item[] Question: Does the research conducted in the paper conform, in every respect, with the NeurIPS Code of Ethics \url{https://neurips.cc/public/EthicsGuidelines}?
    \item[] Answer: \answerYes{} % Replace by \answerYes{}, \answerNo{}, or \answerNA{}.
    \item[] Justification: Nothing to violate in this work.

\item {\bf Broader impacts}
    \item[] Question: Does the paper discuss both potential positive societal impacts and negative societal impacts of the work performed?
    \item[] Answer: \answerYes{} % Replace by \answerYes{}, \answerNo{}, or \answerNA{}.
    \item[] Justification: At Introduction section.
    
\item {\bf Safeguards}
    \item[] Question: Does the paper describe safeguards that have been put in place for responsible release of data or models that have a high risk for misuse (e.g., pre-trained language models, image generators, or scraped datasets)?
    \item[] Answer: \answerNA{} % Replace by \answerYes{}, \answerNo{}, or \answerNA{}.
    \item[] Justification: Not related.

\item {\bf Licenses for existing assets}
    \item[] Question: Are the creators or original owners of assets (e.g., code, data, models), used in the paper, properly credited and are the license and terms of use explicitly mentioned and properly respected?
    \item[] Answer: \answerYes{} % Replace by \answerYes{}, \answerNo{}, or \answerNA{}.
    \item[] Justification: Yes.

\item {\bf New assets}
    \item[] Question: Are new assets introduced in the paper well documented and is the documentation provided alongside the assets?
    \item[] Answer: \answerYes{} % Replace by \answerYes{}, \answerNo{}, or \answerNA{}.
    \item[] Justification: We detailed at main section.

\item {\bf Crowdsourcing and research with human subjects}
    \item[] Question: For crowdsourcing experiments and research with human subjects, does the paper include the full text of instructions given to participants and screenshots, if applicable, as well as details about compensation (if any)? 
    \item[] Answer: \answerNA{} % Replace by \answerYes{}, \answerNo{}, or \answerNA{}.
    \item[] Justification: Not related.

\item {\bf Institutional review board (IRB) approvals or equivalent for research with human subjects}
    \item[] Question: Does the paper describe potential risks incurred by study participants, whether such risks were disclosed to the subjects, and whether Institutional Review Board (IRB) approvals (or an equivalent approval/review based on the requirements of your country or institution) were obtained?
    \item[] Answer: \answerNA{} % Replace by \answerYes{}, \answerNo{}, or \answerNA{}.
    \item[] Justification: Never. not related.

\item {\bf Declaration of LLM usage}
    \item[] Question: Does the paper describe the usage of LLMs if it is an important, original, or non-standard component of the core methods in this research? Note that if the LLM is used only for writing, editing, or formatting purposes and does \emph{not} impact the core methodology, scientific rigor, or originality of the research, declaration is not required.
    %this research? 
    \item[] Answer: \answerYes{} % Replace by \answerYes{}, \answerNo{}, or \answerNA{}.
    \item[] Justification: To polish writing.

\end{enumerate}
\end{document}